\documentclass[letterpaper]{article} 
\usepackage{aaai19}  
\usepackage{times}  
\usepackage{helvet}  
\usepackage{courier}  
\usepackage{url}  
\usepackage{graphicx}  
\usepackage{multirow}  
\usepackage{subfigure}
\usepackage[ruled,linesnumbered]{algorithm2e}
\newtheorem{theorem}{Theorem}
\newtheorem{lemma}{Lemma}
\newcommand*\samethanks[1][\value{footnote}]{\footnotemark[#1]}
\newcommand{\ourmethod}{\emph{Balanced Sparsity}}

\title{Balanced Sparsity for Efficient DNN Inference on GPU}
\author{Zhuliang Yao
\textsuperscript{1,4,}\protect\thanks{Equal contribution.}, 
Shijie Cao
\textsuperscript{2,4,}\protect\samethanks[1], 
Wencong Xiao
\textsuperscript{3,4}, 
Chen Zhang
\textsuperscript{4}, 
Lanshun Nie
\textsuperscript{2}\\
\textsuperscript{1}Tsinghua University
~\textsuperscript{2}Harbin Institute of Technology
~\textsuperscript{3}Beihang University
~\textsuperscript{4}Microsoft Research Asia\\
\{v-zhuyao, v-shicao, v-wencxi, zhac\}@microsoft.com,
nls@hit.edu.cn
}

\begin{document}
\maketitle

\begin{abstract}
In trained deep neural networks, unstructured pruning can reduce redundant weights to lower storage cost. However, it requires the customization of hardwares to speed up practical inference. Another trend accelerates sparse model inference on general-purpose hardwares by adopting coarse-grained sparsity to prune or regularize consecutive weights for efficient computation. But this method often sacrifices model accuracy.
In this paper, we propose a novel fine-grained sparsity approach, \ourmethod{}, to achieve high model accuracy with commercial hardwares efficiently.
Our approach adapts to high parallelism property of GPU, showing incredible potential for sparsity in the widely deployment of deep learning services.
Experiment results show that \ourmethod{} achieves up to 3.1x practical speedup for model inference on GPU, while retains the same high model accuracy as fine-grained sparsity.
\end{abstract}

\section{Introduction} \label{intro}
In the past few years, deep neural network (DNN) has achieved remarkable state-of-the-art results with large-scale network models for many challenging tasks, including computer vision~(CV), natural language processing~(NLP), and speech recognition.
However, recent researches show that the significant redundancy exists in trained model weights, reaching up to 98\% for popular computer vision models \cite{han2015learning,deepcompression}. 
Driven by the great potentials to reduce the model sizes for accelerating DNNs,
a series of work \cite{han2015learning,guo2016dynamic,molchanov2016pruning,lecun1990optimal,engelbrecht2001new} identify and zero out the unimportant weights at a high compression ratio. 
Redundant weight pruning methods keep model accuracy and often benefit DNN models in cost-effective service deployment with much fewer resources.

Despite a significant reduction in operative weights, the fine-grained sparsity can only save storage costs, but hardly speed up inference due to the fragmented unstructured weights in pruned models.
The irregularity and random distribution in weight matrices poorly fit current general purpose accelerators (i.e. GPU), which often advocate highly parallel computing characteristic.
The speedup could be negative when the sparsity ratio is quite low and only a sparsity ratio higher than 95\% can lead to speedup.\cite{wen2016learning,DBLP2018WangZWH}
Therefore, customized hardwares~\cite{eie,ese,scnn} are required for the widely deployment of model pruning.

Another research work chooses to maintain a dense structure during pruning.
More specifically, pruning granularity often incorporates with neural network semantics in convolution neural network (CNN) structures, e.g., filter and channel~\cite{li2016pruning,he2017channel} and recurrent neural network (RNN) states, e.g., cell and gate~\cite{wen2018learning}.
With coarse-grained DNN component pruned, the remaining parameters are still in a compact structure which is a quite hardware-friendly feature and make practical acceleration more possible.
However, despite the notable speedup observed, the pruned models usually compromise accuracy.

\begin{figure}[t]
\centering
\includegraphics[width=0.93\linewidth]{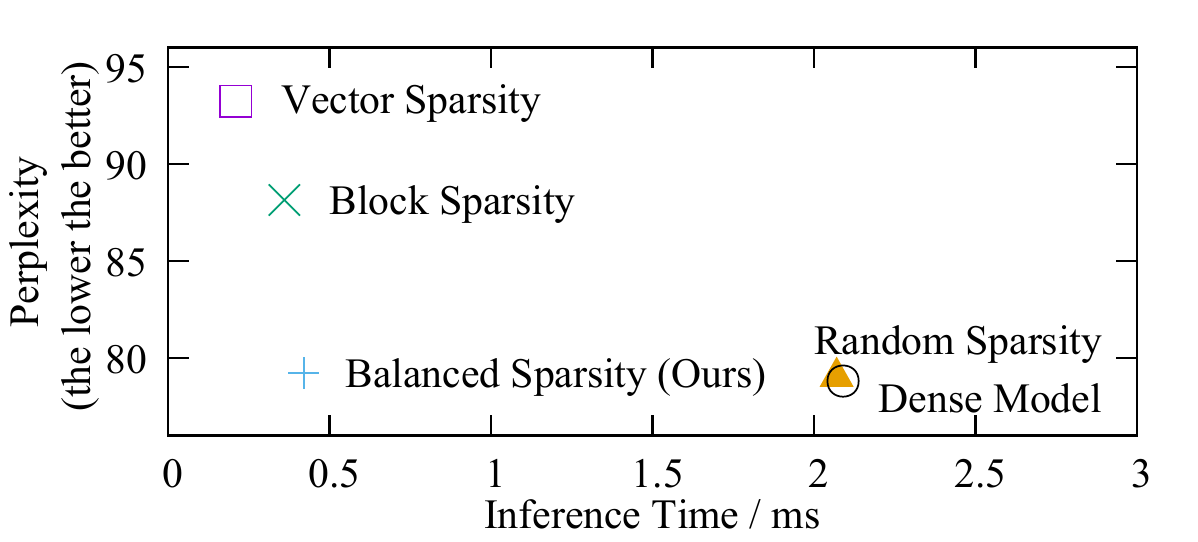}
\caption{Perplexity and Inference Time trade-off of different sparsity patterns on the PTB dataset~\cite{ptb}.
All the methods prune the same pre-trained LSTM model with single 1500-hidden-units cell to reach 90\% sparsity.
}
\label{fig:intro}
\end{figure}

Figure~\ref{fig:intro} shows the model accuracy and inference time trade-off for pruning a trained LSTM model with different sparsity patterns. 
The random sparsity~\cite{han2015learning} approach is poor in inference speed while almost achieving the same accuracy as the original dense model.
On the contrary, coarse-grained sparsity patterns, i.e both vector sparsity~\cite{mao2017exploring} and block sparsity~\cite{narang2017block} fit GPU architecture for matrix operation acceleration, however, losing in model accuracy.

To leverage sparsity for inference acceleration on GPUs while retaining model accuracy, we thereby propose a novel sparsity pattern, \ourmethod{}.
\ourmethod{} aims at pruning model weights in a balanced structure. 
Instead of pruning a weight matrix in a monolithic way, we partition the weight matrix and perform independent pruning in sub-matrices.
We conduct a set of experiments on typical neural networks to show the performance of our method, focusing on model accuracy and inference time.
For accuracy, our experiments on three typical CV, NLP, and Speech tasks show that, we achieve less than 0.2\% accuracy difference comparing with fine-grained random sparsity.
For inference time, our benchmark result shows that, we achieve almost ideal performance speedup on GPU for matrix multiplication under the sparsity ratio ranging from 50\% to 97\%. On PTB dataset, our \ourmethod{} obtains coarse-grained level speedup and keeps fine-grained level accuracy (Figure~\ref{fig:intro}).
Besides, a series of detailed measurements on typical networks, including VGG-16 net, LSTM model, and CTC model, show that \ourmethod{} achieves 1.4x to 3.1x practical speedup in GPU inference.

Overall, we make three contributions in this paper:
\begin{itemize}
\item We propose a new sparsity pattern \ourmethod{} and the corresponding pruning method that can both maintain model accuracy and achieve significant practical acceleration.
\item We provide a matrix operation implementation based on the special architecture design inside GPU.
\item Our \ourmethod{} achieves the state-of-the-art practical speedup while keeps the same high model accuracy as both dense model and random sparsity approach.
\end{itemize}

\section{Related Work}

\subsection{Fine-grained Sparsity}
The redundancy of neural network is well recognized by LeCun et al. \cite{lecun1990optimal} since 1990s.
Recent years, fine-grained weight pruning approach removes over 90\% of weight parameters in popular CV models, significantly reducing the model size for model deployment and inference services.
Iterative pruning~\cite{han2015learning} is firstly introduced, which prunes individual weights below a monotonically increasing threshold value and then retrains the remaining weights iteratively.
Meanwhile, its capability to retain model accuracy is justified on a wide range of popular neural network models of CNN~\cite{guo2016dynamic,aghasi2017net,liu2018efficient} and RNN~\cite{giles1994pruning,lin2017runtime}.
However, redundancy-orient pruning introduces irregularity in model.
Custom hardwares~\cite{eie,ese,scnn} are essential to speedup the computing for fragmented random data accesses, which limit the deployment of sparse DNNs.

\subsection{Coarse-grained Sparsity}
Recent research observes the irregularity challenge in model sparsity and falls back to consider the support for general purposed processors.
Not only weight importance but also neural network semantics are jointly considered in model pruning. 
The goal is to generate a sparse output while keeping dense sub-structures, therefore pruning is usually applied on coarse-grained model component.
Filter and channel level sparsity for CNN~\cite{li2016pruning,neklyudov2017structured,wen2016learning}, 
gate and cell state sparsity for RNN~\cite{wen2018learning}, 
low rank approximation~\cite{jaderberg2014speeding,liu2015sparse}, 
and block sparsity~\cite{narang2017block} are several sparsity patterns in which model structure is fully considered.
As pointing out by~\cite{mao2017exploring,zhu2017prune}, the coarse-grained sparsity 
benefits computation-intensive accelerators (e.g. GPU), however, causes prominent accuracy penalty comparing with fine-grained approaches.
These methods~\cite{mao2017exploring,narang2017block} modify the iterative pruning method to apply in consecutive weight blocks. They pick the maximum magnitude or the average magnitude of the weights within one block as a representative for the entire block. A monotonically increasing threshold is adopted also.

\section{Methodology}  \label{sparsity}

Neural network pruning methods bring a restricted freedom to define the sparsity structure (e.g. hardware friendly sparsity) in weight matrices.
More regular sparsity structure can increase hardware efficiency, but is also easier to destroy the original distribution of weight matrices which may hurt model accuracy significantly.
Ideally, a good sparsity structure should {} balance model accuracy and hardware efficiency.

Our proposed sparsity pattern, \ourmethod{}, achieves both high model accuracy and high hardware efficiency.
In this section, we first introduce the \ourmethod{} sparsity pattern and the balance-aware iterative pruning algorithm to induce \ourmethod{}.
Then, we use a mathematical way to prove that the influence on model accuracy is limited.
Finally, we present an efficient GPU implementation for our \ourmethod{}.

\subsection{Balanced Sparsity}  \label{sec:sparsity}
\begin{figure}[h]
	\centering
    \includegraphics[width=\linewidth]{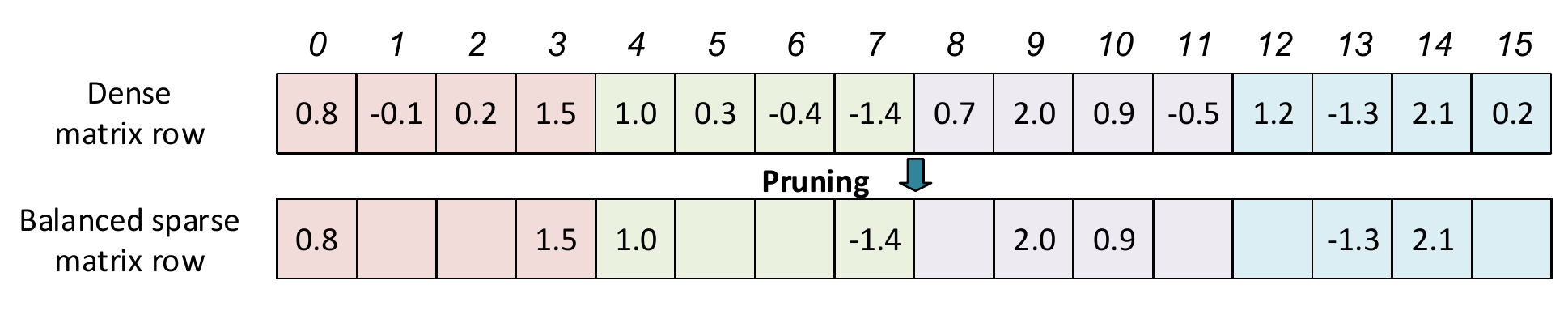}
    \caption{An example of pruning a dense matrix row to a \ourmethod{} matrix row.}
	\label{fig:bs}
\end{figure}

To maintain high model accuracy and achieve efficient GPU acceleration, we propose a novel fine-grained sparsity, called \ourmethod{}. 
For weight matrices represented in \ourmethod{}, each matrix row is split into multiple equal-sized blocks and each block has the same number of non-zero weights. 
Figure \ref{fig:bs} shows an example of a block-balanced sparse matrix row pruned from a dense matrix row. In this example, the matrix row is split into 4 blocks, and each block has a sparsity of 50\%. The balance range, i.e the length of each block, is 4. The same split method and sparsity apply to other rows in the weight matrix.

The intuitions of designing the \ourmethod{} are: 1) the block partition with balanced computation work load for each block naturally fit GPUs with high practical parallelism. 2) the random distribution of non zero weights inside a block adds very few constraints on the sparsity structure and may not affect model accuracy. 

\subsection{Balance-aware Iterative Pruning}  \label{sec:pruning}
We introduce a balance-aware iterative pruning method to induce \ourmethod{} to weight matrices.
For CNNs, the weights of all kernels in one convolution layer are considered as one weight matrix.
Previous pruning methods usually adopt a monotonically increasing threshold value to zero out the weights less than this threshold. Those methods do not consider the distribution of non-zero values. 

We use an expected sparsity instead of a threshold value to prune weights, which guarantees a balanced distribution of non-zero weights among block partitions during pruning iterations.
Algorithm \ref{alg:pruning} illustrates our balance-aware iterative pruning method.
In each pruning iteration, the pruning algorithm sorts the weights in each block by their absolute magnitude and then zeros out a fraction of weights with smallest absolute magnitudes under the threshold percentage.
This threshold percentage is gradually increased from 0 to the target sparsity while the increase rate decreases with pruning iteration. Figure \ref{fig:bs} illustrates a balance-aware pruning iteration with a threshold sparsity of 50\%.  

\begin{algorithm}[h]
\caption{Balance-aware Iterative Pruning}
\label{alg:pruning}

\KwIn{The matrix to be pruned, $M$\;
~~~~~~~~~~~~The number of blocks per row, $BlockNum$\;
~~~~~~~~~~~~The expected sparsity, $Sparsity$\;
}
\KwOut{The pruned matrix, $M_p$;}

\For{$M_i \in M.rows$}{
Divide $M_i$ into $block_{i,j}$ ($j=1~to~BlockNum$)\;
}
$tmp_{sparsity} = 0$\;
\While{$tmp_{sparsity} < Sparsity$}{
$tmp_{sparsity} = GraduallyIncrease(tmp_{sparsity})$\;
\For{$block_{i,j} \in M$}{
Sort elements and calculate the block internal threshold $ T_{i,j} $ based on $tmp_{sparsity}$\;
\For {each $element \in block_{i,j}$}{
prune $element$ $\textbf{if}$ {$|element| < T$}\;
}
}
}
return the pruned matrix, $M_p$;
\end{algorithm}

In our method, pruning followed by a retraining is one iteration, which is also defined in previous methods~\cite{han2015learning,mao2017exploring,narang2017block}.
For multi-layer network like VGG-16 net, we adopt a straightforward strategy which separates the whole net into layers, then prune all those convolutional layers and FC layers one by one.

\subsection{Asymptotic Analysis}		\label{sec:math}
We prove that the influence of our \ourmethod{} on model accuracy is very slight, by theoretically showing that the differences between random sparsity~\cite{han2015learning} and our method are negligible for practical situations.
To compare the similarities and differences between these two methods, we perform a theoretical analysis on a fully-connected layer:
\begin{equation}
Y=W^{(0)}\cdot X+B,
\end{equation}
where $W^{(0)}$ is an $M\times N$ matrix, $X$ is an $N$-dimensional vector of input features, $B$ is an $M$-dimensional vector of bias term, and $Y$denotes the output of this fully-connected layer. For ease of elaboration, we assume that the bias vector $B$ is a zero vector here.

Similar to many prior works
\cite{hernandez2015probabilistic,blundell2015weight,salimans2016weight}, we specify an independent Gaussian priors distribution $\mathcal{N}(0,\sigma_w^2)$ for each element $w$ in $W^{(0)}$ and another $\mathcal{N}(0,\sigma_x^2)$ for each element $x$ in input $X$. Then the output difference between sparse and dense FC-layer can be denoted as
\begin{equation}
	Z^{(i)}=W^{(i)}\cdot X - W^{(0)}\cdot X=dW^{(i)}\cdot X,\quad \forall~i \in \{1,2\}
\end{equation}
where $W^{(1)}$ is the matrix pruned with random sparsity and $W^{(2)}$ is the matrix pruned with \ourmethod{}.

Firstly, we defined a function $H\left(k\right)$ as follows,
\begin{equation}
H\left(k\right) = \frac{k\left(MN-k\right)}{(MN)^3\cdot \left[ f\left( F^{-1}\left( \frac{k}{MN} \right) \right) \right]^2},
\end{equation}
where $f$ and $F$ are probability density function and cumulative distribution function of $W^{(0)}$'s Gaussian distribution, $F^{-1}$ denotes the quantile function associated with $F$.
\begin{lemma}
\label{lemma_char}
The characteristic functions of the variable $z$'s distributions in $Z^{(i)}~$,$\forall~i \in \{1,2\}$, are
\begin{equation}\label{equ_result1}
\Phi _{Z^{(1)}}\left( t\right )=\frac{\sigma_x}{r^{(1)}} \left( 1+t^2\right)^{-\frac{1}{2}}\sum_{i=1}^{r^{(1)} }H\left({i}\right)
\end{equation}
and
\begin{equation}\label{equ_result2}
\Phi _{Z^{(2)}}\left( t\right )=\frac{\sigma_x}{r^{(1)}} \left( 1+t^2\right)^{-\frac{1}{2}}\sum_{i=1}^{r^{(1)} }H\left(\left \lceil \frac{i}{MK}  \right \rceil\times MK\right),
\end{equation}
where $K$ is the number of balance range, $r^{(1)}=MK \cdot r^{(2)}$ is the total number of pruned elements.
\end{lemma}

With the help of Lemma \ref{lemma_char}, we get the following theorem:
\begin{theorem}
\label{theorem_dist}
The means of the variable $z$'s distributions in $Z^{(i)}~$,$\forall~i \in \{1,2\}$, are 
\begin{equation}
Mean_{Z^{(1)}}\left(z\right) = Mean_{Z^{(2)}}\left(z\right) = 0.
\end{equation}
The variances the variable $z$'s distributions in $Z^{(i)}~$,$~\forall~i \in \{1,2\}$, are 
\begin{equation}
Var_{Z^{(1)}}\left(z\right)=\frac{\sigma_x}{r^{(1)}}\sum_{i=1}^{r^{(1)}}  H\left(i\right)
\end{equation}
and
\begin{equation}
Var_{Z^{(2)}}\left(z\right)=\frac{\sigma_x}{r^{(1)}}\sum_{i=1}^{r^{(1)}} H\left(\left \lceil \frac{i}{MK}  \right \rceil\times MK\right). 
\end{equation}
\end{theorem}

As showed in equations (\ref{equ_result1}) and (\ref{equ_result2}), $\Phi _Z^{(1)}\left( t\right )$ and $\Phi _Z^{(2)}\left( t\right )$ have similar formulation. The mean values of random sparsity and our purposed \ourmethod{} are both equal to zero. And the difference between their variances can be regarded as a limited quantization error (i.e., $i$ v.s. $\left \lceil \frac{i}{MK}  \right \rceil\times MK$ ).
The analysis result is consistent to what we observe in real workloads as visualized in experiments. Please refer to \url{https://github.com/Howal/balanced-sparsity/blob/master/appendix-aaai19.pdf} for proof.

\subsection{Efficient GPU Implementation}  \label{sec:gpu} 

We now introduce our efficient GPU library of matrix multiplication for balanced sparse matrices.

\begin{figure}[h]
	\centering
    \includegraphics[width=0.9\linewidth]{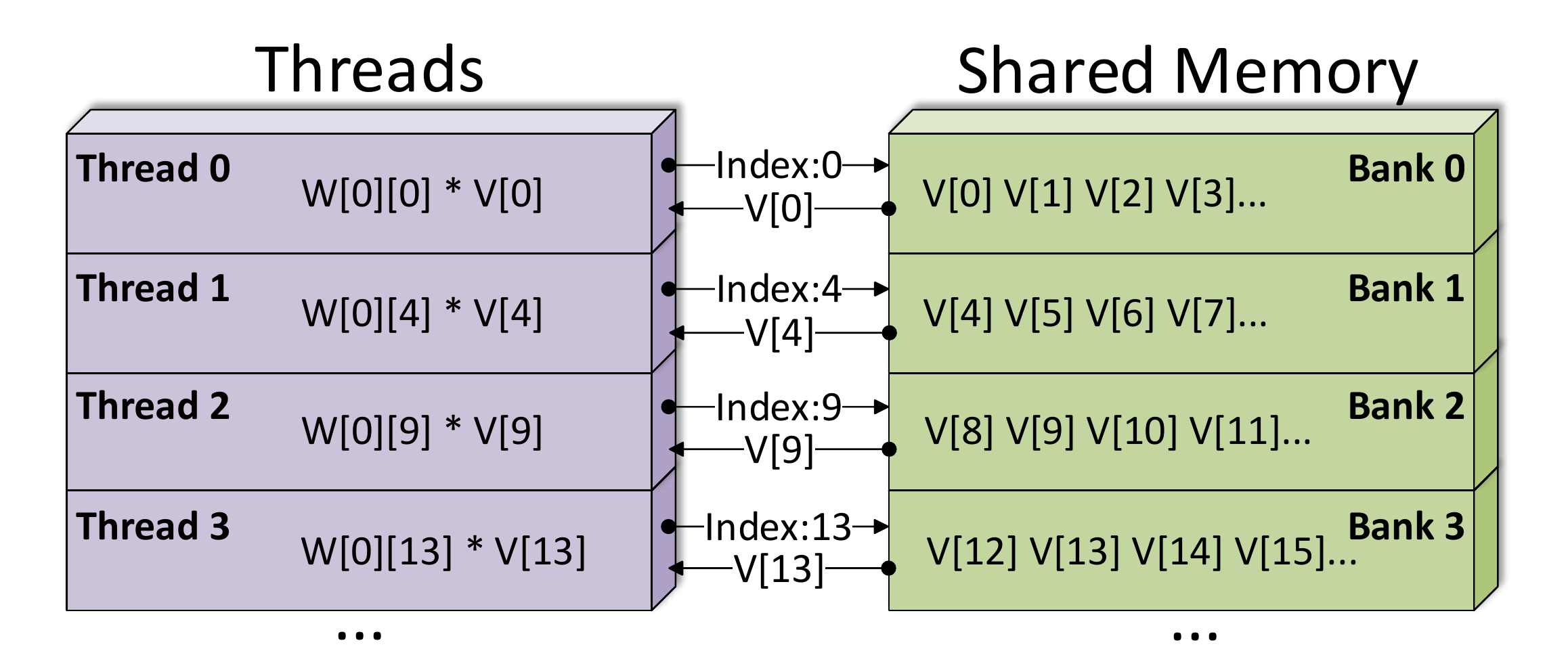}
    \caption{Parallelizing threads for efficient sparse matrix multiplication. Shared memory supplies V[0], V[4], V[9], V[13] simultaneously according to the indexes.}
	\label{fig:bank}
\end{figure}

Our implementation first utilizes the block partition as a workload partition for GPUs to achieve high practical parallelism.
Modern GPUs contain massive cores that can support thousands of threads running simultaneously. 
In our case, the multiplication and accumulation operations in one block partition are assigned to a single thread.
The same number of non-zero values in each block partition can further increase the GPU efficiency because it makes the workloads between threads balance.

Sparse matrices after pruning lose the regular structure of dense matrices which results in irregular memory accesses in sparse matrix multiplication. 
Running massive threads in parallel causes concurrent random memory access problem. Improper handling of random memory accesses from various threads could stall the thread execution and decrease the performance significantly.

In order to overcome the challenge in random memory accesses, our implementation takes advantage of the shared memory in GPU architecture to support concurrent random accesses. 
In GPU architecture, a chunk of shared memory is visible to a fixed number of threads.
To achieve high memory bandwidth for concurrent accesses, shared memory is divided into equally sized memory modules, which is called \emph{banks} that can be accessed independently and simultaneously. Therefore, any memory load or store of $n$ addresses that spans $n$ distinct memory banks can be serviced simultaneously, yielding an effective bandwidth that is $n$ times as high as the bandwidth of a single bank. 
In Figure \ref{fig:bank}, we use the balanced sparse matrix in Figure \ref{fig:bs} as an example to shows how to parallelize the threads for sparse matrix multiplication. The dense vector to be multiplied is rearranged and stored in shared memory to avoid bank conflicts.

\section{Experiments}   \label{Experiments}

In this section, we compare \ourmethod{} against the dense model baseline, random sparsity\cite{han2015learning}, block sparsity~\cite{narang2017block}, and vector sparsity~\cite{mao2017exploring} for model accuracy.
For the GPU inference performance test, we use different highly optimized libraries for different sparsity patterns.
The baseline of dense matrices is tested with the cuBLAS library.
For random sparse matrices, we use the cuSPARSE library.
For block sparse matrices, we use an open sourced GPU library \cite{blocksparsecode}, which is highly optimized for matrix multiplication of block sparse matrices on GPU. 
For balanced sparse matrices, we use our own GPU implementation as described above.
Vector sparsity is not evaluated here, because there is no available GPU implementation as far as we know.

The experiments are divided into three parts. Firstly, we test our GPU implementation on a matrix multiplication benchmark. Then we apply our sparsity approach to multiple wide-used deep learning workloads, covering CV, NLP, and Speech. Finally, we investigate the feature of our sparsity pattern in further detail by visualizing the weight map and tuning the hyper-parameter, balance range. All the experiments in this section are done with a batch size of 1, the block number per row of our method is 32, and the block size of block sparsity is $8*8$, unless explicitly stated.

\subsection{Benchmark}  \label{bench}

In order to show the hardware efficiency of our proposed \ourmethod{}, we conduct a benchmark to compare the inference time of a matrix multiplication among all existing sparsity patterns. This benchmark uses a matrix size of 16384 $\times$ 8196. 

\begin{figure}[t]
\centering
\subfigure[batchsize = 1]{
\begin{minipage}[b]{\linewidth}
\centering
\includegraphics[width=0.9\linewidth]{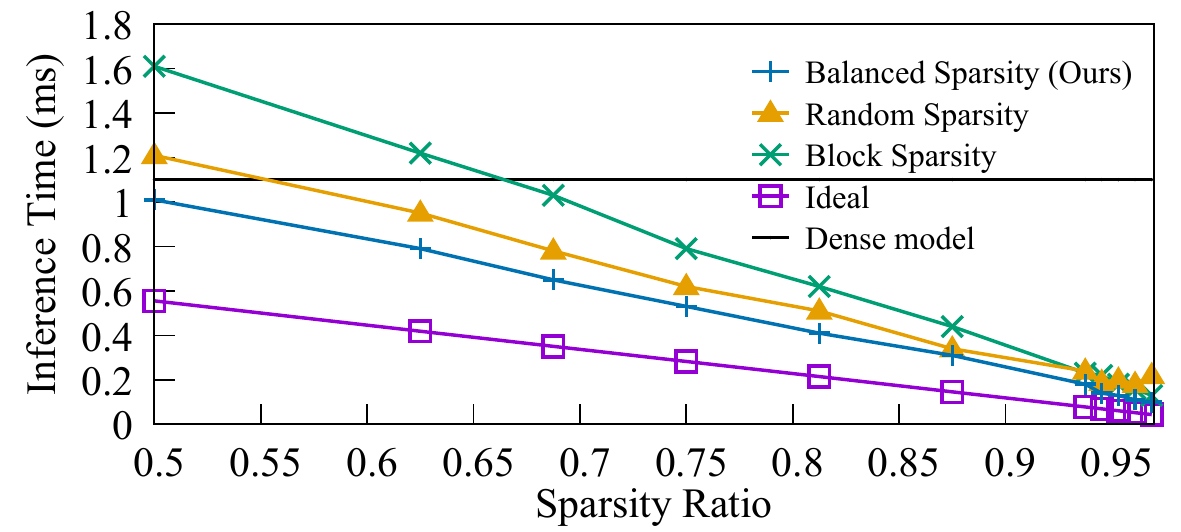}
\end{minipage}
}
\subfigure[batchsize = 8]{
\begin{minipage}[b]{\linewidth}
\centering
\includegraphics[width=0.9\linewidth]{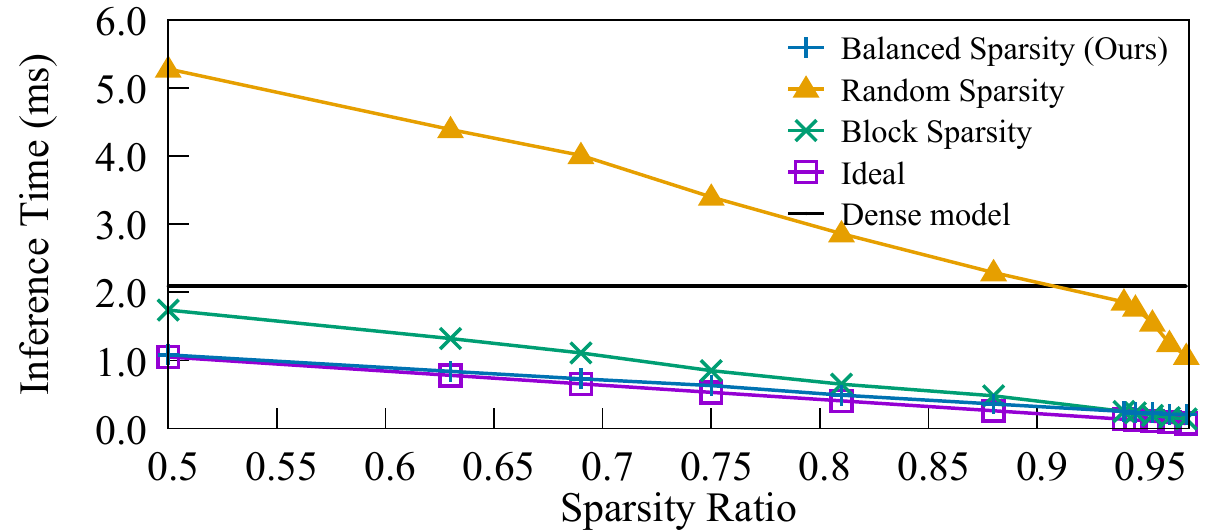}
\end{minipage}
}
 \caption{Inference time benchmark comparisons of various sparsity patterns.} \label{fig:benchmark}
\end{figure}

Figure \ref{fig:benchmark} shows the speedup of \ourmethod{} with our GPU implementation.
In this benchmark of matrix multiplication, our method outperforms other sparsity patterns. When $batch size = 1$, there is still a gap between our method and idea time, because the main benchmark bottleneck of this setting is the communication inside GPU. This disadvantage can be overcome by hiding the I/O time in more batches.
For $batch size = 8$ case, our method almost reaches the ideal inference time brought by skipping unnecessary computation. 
The ideal inference time ($i\_time$) is calculated by the following equation:
\begin{equation}
i\_time = (d\_time - o\_time) * (1-sparsity) + o\_time
\end{equation}
where the $d\_time$ denotes the inference time of a dense matrix running on cuBLAS, the $o\_time$ denotes the time overhead of launching an execution kernel on GPU. Here we take 10us as $o\_time$ which is a widely used number~\cite{chu2016cuda}.

Notice that using cuSPARSE for sparse computation can achieve speedup only if the sparsity ratio is higher than around 91\%, while our method is always faster than the dense baseline.

\subsection{Real Workloads} \label{workloads}

In this subsection, we apply our balanced sparsity pattern to vision, language, and speech tasks.
We compare the compression rate (i.e. achievable sparsity) of our balanced sparsity with other four alternatives, including dense model baseline, random sparsity, block sparsity, and vector sparsity.
Random sparsity performs pruning in each independent weight matrix. 
Block sparsity treats a consecutive block of parameters as a pruning unit. If a pruning decision is made, the whole block weights will be removed. 
Vector sparsity means to consider a whole row or column in a weight matrix as a basic pruning unit.

In our pruning experiments, we apply the same hyper-parameters and fine-tune techniques to various sparsity patterns.
During pruning, if the model accuracy drops significantly and cannot recover via retraining, we withdraw this pruning iteration and stop the pruning procedure.
For practical speedup, we compare our GPU implementation with other available GPU implementations for dense model, random sparse model, and block sparse model. 

\paragraph{VGG-16 on ImageNet} For the vision task, we use VGG-16 network~\cite{vgg} on ImageNet ILSVRC-2012 dataset~\cite{krizhevsky2012imagenet} to evaluate the compression rate and practical speedup. VGG-16 is a well-known network architecture which contains 13 convolutional layers and 3 FC layers, while the dataset has 1.2M training examples and 50k validation examples.

We use random sparsity, block sparsity, and balanced sparsity to prune both convolutional and fully-connected layers of a pre-trained VGG-16 model, respectively.
Then we evaluate the inference time of those pruned models with their customized GPU implementations. 
One popular implementation of convolution operation is using im2col that converts convolution operation to matrix-matrix multiplication ~\cite{im2col}. The operation of a fully-connected layer is matrix-vector multiplication.

\begin{figure}[t]
	\centering
	\includegraphics[width=0.9\linewidth]{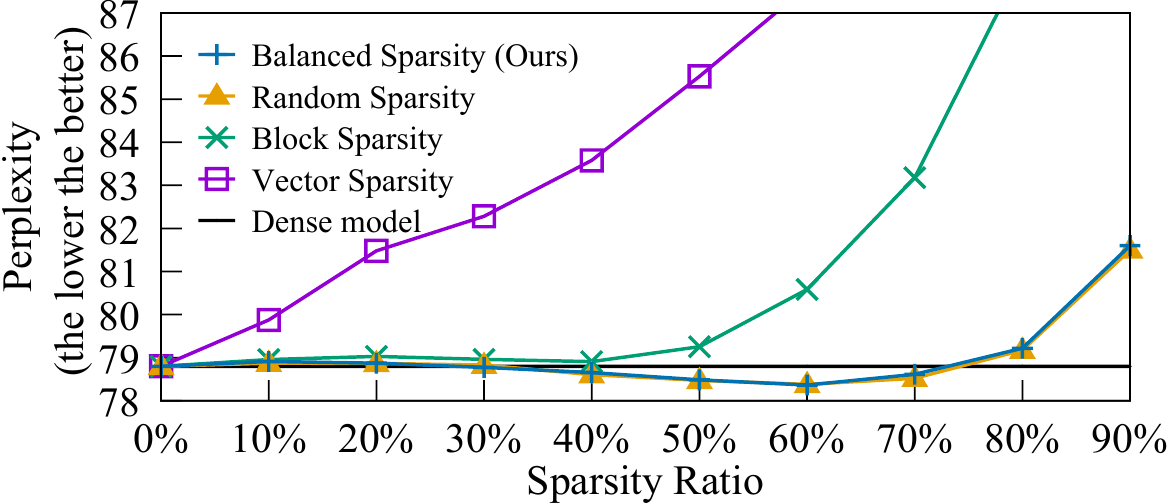}
	\caption{Sparsity-Perplexity curves of various sparsity patterns on PTB dataset.}
	\label{fig:ptb}
\end{figure}

\begin{table}[t]
\centering
\resizebox{0.9\linewidth}{!}{
\begin{tabular}{c|c|c|c}
\hline
\multicolumn{2}{c|}{\begin{tabular}[c]{@{}c@{}}Language Model /\\ PTB\end{tabular}}                                            & \begin{tabular}[c]{@{}c@{}}Inference\\ Time / us\end{tabular} & Sparsity      \\ \hline\hline
\multirow{4}{*}{\begin{tabular}[c]{@{}c@{}}Sparsity\\ Patterns\end{tabular}} & Dense Model                                     & 294.1                                                         & 0\%           \\ \cline{2-2}
                                                                             & Random Sparsity                                 & 370.9                                                         & \textbf{80\%} \\ \cline{2-2}
                                                                             & Block Sparsity                                  & 326.3                                                         & 40\%$^*$          \\ \cline{2-2}
                                                                             & \multicolumn{1}{l|}{\textbf{Balanced Sparsity}} & \textbf{120.2}                                                & \textbf{80\%} \\ \hline
\end{tabular}
}
\caption{Inference time comparisons of various sparsity patterns on PTB dataset. Our methods outperforms all the other methods. $^*$Block Sparsity could only reach a sparsity ratio of 40\% without hurting the performance.}
\label{ptbperf}
\end{table}

Table \ref{cnn-perf} reports the layer-wise results and the whole model result. All these three methods as well as the dense model baseline achieve similar top-5 accuracy of 90.3\%, however, under different sparsity ratios. In terms of compression rate, both random sparsity and our balanced sparsity can compress the VGG-16 model with more than 12x, but block sparsity can only compress the model with less than 4x. 
Our GPU implementation for balanced sparsity also achieves the best practical speedup, which is 6x faster than random sparsity.

\begin{table*}[htbp]
\centering
\resizebox{0.8\linewidth}{!}{
\begin{tabular}{c|c|c|c|c|c|c|cc}
\hline
\multirow{2}{*}{} & \multicolumn{2}{c|}{Dense Model}                                                      & \multicolumn{2}{c|}{Random Sparsity}                                                  & \multicolumn{2}{c|}{Block Sparsity}                                                   & \multicolumn{2}{c}{Balanced Sparsity}                                                                            \\ \cline{2-9} 
                  & \begin{tabular}[c]{@{}c@{}}Inference\\ Time \textbackslash us\end{tabular} & Sparsity & \begin{tabular}[c]{@{}c@{}}Inference\\ Time \textbackslash us\end{tabular} & Sparsity & \begin{tabular}[c]{@{}c@{}}Inference\\ Time \textbackslash us\end{tabular} & Sparsity & \multicolumn{1}{c|}{\begin{tabular}[c]{@{}c@{}}Inference\\ Time \textbackslash us\end{tabular}} & Sparsity       \\ \hline\hline
conv1\_1          & 144.0                                                                      & -        & 714.7                                                                      & 42\%     & 78.3                                                                       & 31\%     & \multicolumn{1}{c|}{254.7}                                                                      & 34\%           \\
conv1\_2          & 612.5                                                                      & -        & 2578.0                                                                     & 88\%     & 949.4                                                                      & 56\%     & \multicolumn{1}{c|}{1018.4}                                                                     & 68\%           \\
conv2\_1          & 393.5                                                                      & -        & 1842.5                                                                     & 70\%     & 356.2                                                                      & 41\%     & \multicolumn{1}{c|}{474.4}                                                                      & 65\%           \\
conv2\_2          & 588.2                                                                      & -        & 4640.0                                                                     & 71\%     & 639.9                                                                      & 38\%     & \multicolumn{1}{c|}{557.0}                                                                      & 71\%           \\
conv3\_1          & 305.0                                                                      & -        & 2668.6                                                                     & 57\%     & 286.2                                                                      & 30\%     & \multicolumn{1}{c|}{371.4}                                                                      & 45\%           \\
conv3\_2          & 584.4                                                                      & -        & 3768.9                                                                     & 84\%     & 362.6                                                                      & 56\%     & \multicolumn{1}{c|}{396.5}                                                                      & 79\%           \\
conv3\_3          & 584.4                                                                      & -        & 4257.4                                                                     & 71\%     & 490.3                                                                      & 35\%     & \multicolumn{1}{c|}{355.7}                                                                      & 88\%           \\
conv4\_1          & 333.3                                                                      & -        & 2005.3                                                                     & 79\%     & 237.8                                                                      & 41\%     & \multicolumn{1}{c|}{295.4}                                                                      & 86\%           \\
conv4\_2          & 623.0                                                                      & -        & 3196.0                                                                     & 86\%     & 316.6                                                                      & 57\%     & \multicolumn{1}{c|}{366.2}                                                                      & 91\%           \\
conv4\_3          & 623.0                                                                      & -        & 3205.9                                                                     & 85\%     & 500.5                                                                      & 38\%     & \multicolumn{1}{c|}{396.5}                                                                      & 88\%           \\
conv5\_1          & 211.0                                                                      & -        & 920.1                                                                      & 88\%     & 170.7                                                                      & 41\%     & \multicolumn{1}{c|}{129.9}                                                                      & 86\%           \\
conv5\_2          & 211.0                                                                      & -        & 926.3                                                                      & 91\%     & 132.9                                                                      & 52\%     & \multicolumn{1}{c|}{126.4}                                                                      & 90\%           \\
conv5\_3          & 211.0                                                                      & -        & 1053.6                                                                     & 89\%     & 163.8                                                                      & 36\%     & \multicolumn{1}{c|}{110.2}                                                                      & 95\%           \\
fc6               & 979.9                                                                      & -        & 1084.6                                                                     & 93\%     & 841.8                                                                      & 75\%     & \multicolumn{1}{c|}{231.1}                                                                      & 93\%           \\
fc7               & 265.5                                                                      & -        & 251.0                                                                      & 93\%     & 238.6                                                                      & 75\%     & \multicolumn{1}{c|}{70.3}                                                                       & 93\%           \\
fc8               & 144.5                                                                      & -        & 294.5                                                                      & 75\%     & 120.6                                                                      & 60\%     & \multicolumn{1}{c|}{58.9}                                                                       & 75\%           \\ \hline \hline
Total$^*$             & 6814.141                                                                   & -        & 33407.4                                                                    & 91.8\%    & 5886.1                                                                     & 71.7\%   & \multicolumn{1}{c|}{\textbf{5213.0}}                                                            & \textbf{92.0\%} \\ \hline
\end{tabular}
}
\caption{Inference time and sparsity comparisons of various sparsity patterns on VGG-16. Our balanced sparsity and customized GPU implementation achieve the best compression rate and practical speedup. $^*$The time cost of other layers in VGG-16, such as pooling and batch normalization, is about 230us, which is less than 3\% of entire inference time.}
\label{cnn-perf}
\end{table*}

\paragraph{LSTM on PTB}  \label{sec:lstm_exp}

In the experiment of PTB dataset~\cite{ptb}, we adopts a 2-layer LSTM language model with LSTM hidden layer size of 1500. 
We compare \ourmethod{} with other sparsity patterns by measuring the final pruned model perplexity, a metric to quantify language model quality (the lower the better).

Figure~\ref{fig:ptb} shows perplexity results under different sparsity patterns.
This figure shows that the perplexity curve of our balanced sparsity is very close to the perplexity curve of random sparsity. Both random sparsity and our balanced sparsity can preserve the perplexity until 80\% of weights are pruned. These two patterns achieve even slightly better model quality, compared to the original one even around 60\% sparsity.
The perplexity of vector sparsity starts to increase significantly at a very low sparsity ratio. 
The perplexity of block sparsity starts to increase at a sparsity of 40\%. 
In summary, our balanced sparsity has almost the same efficacy as random sparsity and outperforms both vector sparsity and block sparsity in terms of achievable accuracy and sparsity during pruning. 

We also compare the inference time of our balanced sparsity with dense baseline, random sparsity, and block sparsity.  
Table \ref{ptbperf} shows the speedup results. 
For the PTB LSTM model, our GPU implementation for balanced sparsity achieves 3.1x speedup compared to the random sparse model running on cuSPARSE, 2.7x speedup compared to the block sparse model running on block sparse library, 2.5x speedup compared to the baseline dense model running on cuBLAS.

\begin{figure}[!tbp]
\centering
\includegraphics[width=0.9\linewidth]{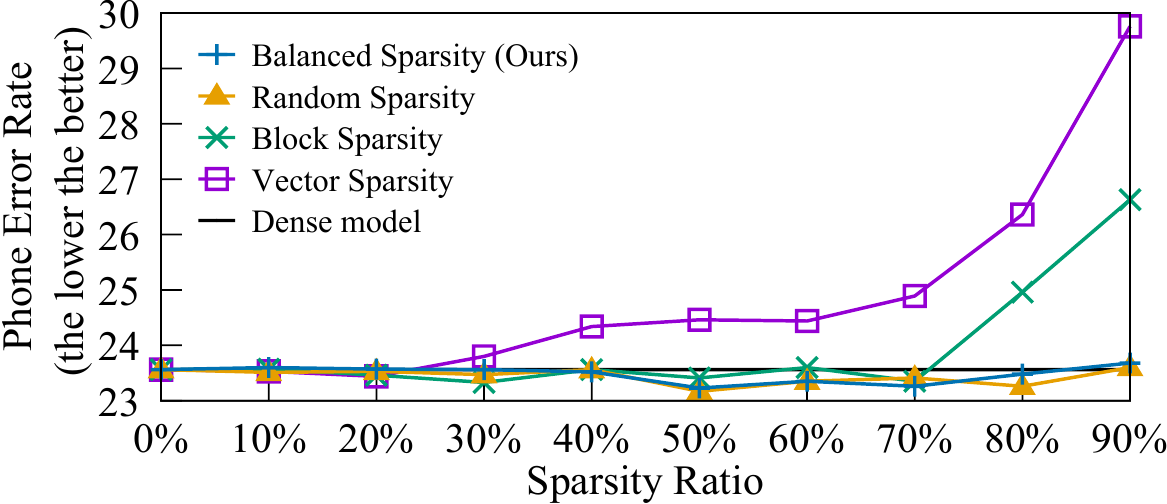}
\caption{Sparsity - Phone Error Rate curves of various sparsity patterns on TIMIT dataset. }
\label{fig:timit}
\end{figure}

\paragraph{CTC on TIMIT}

We further examine our \ourmethod{} on the TIMIT dataset, which is a read speech benchmark and especially designed for acoustic-phonetic studies. A CTC (connectionist temporal classification) model ~\cite{graves2006ctc} is used here, which mainly contains a Bi-LSTM (Bidirectional Long Short-Term Memory) cell with a hidden size of 1024. The settings of different sparsity patterns are the same as mentioned in previous subsection.

\begin{table}[!tbp]
\centering
\resizebox{0.9\linewidth}{!}{
\begin{tabular}{c|c|c|c}
\hline
\multicolumn{2}{c|}{\begin{tabular}[c]{@{}c@{}}Speech Recognition /\\ TIMIT\end{tabular}}                                      & \begin{tabular}[c]{@{}c@{}}Inference\\ Time / us\end{tabular} & Sparsity        \\ \hline\hline
\multirow{4}{*}{\begin{tabular}[c]{@{}c@{}}Sparsity\\ Patterns\end{tabular}} & Dense Model                                     & 117.9                                                         & 0\%             \\ \cline{2-2}
                                                                             & Random Sparsity                                 & 190.5                                                         & \textbf{87.5\%} \\ \cline{2-2}
                                                                             & Block Sparsity                                  & 212.8                                                         & 70\%$^*$            \\ \cline{2-2}
                                                                             & \multicolumn{1}{l|}{\textbf{Balanced Sparsity}} & \textbf{83.9}                                                 & \textbf{87.5\%} \\ \hline
\end{tabular}
}
\caption{Inference time comparisons of various sparsity patterns on TIMIT dataset.
$^*$Notice that the sparsity percentage is chosen based on the accuracy experiment in Figure~\ref{fig:timit}.
Block Sparsity could only reach a sparsity ratio of 70\% without hurting the performance.
}
\label{timitperf}
\end{table}

\begin{figure*}[!t]
	\centering
		\includegraphics[width=0.85\linewidth]{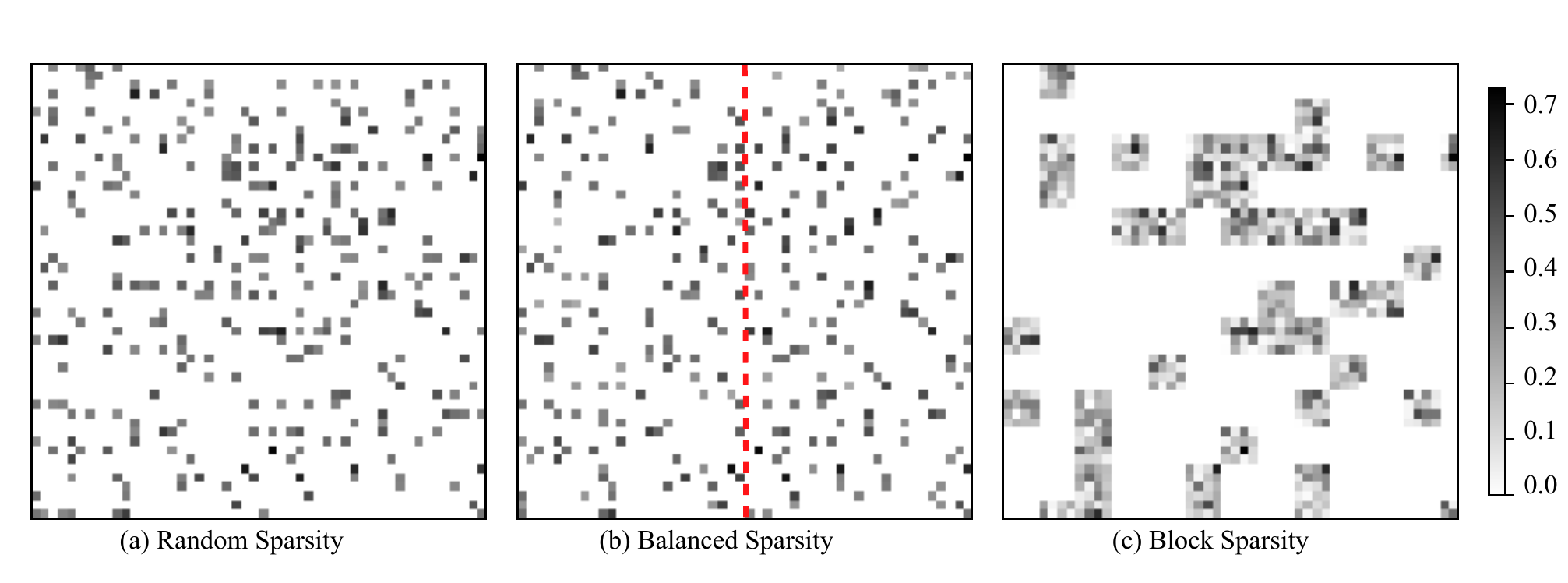}
		\caption{Weight map visualizations after applying random sparsity, \ourmethod{}, and block sparsity (sparsity = 90\%). In (b), each row contains two block partitions (i.e., left side and right side of the dotted line).}
	\label{fig:weights}
\end{figure*}

For the TIMIT Bi-LSTM model, Figure~\ref{fig:timit} shows the perplexity results under different sparsity patterns and Table~\ref{timitperf} shows the inference time of different sparsity patterns. We get the same conclusions as the experiment of PTB LSTM model. In terms of pruning efficacy, our balanced sparsity is similar to random sparsity and outperforms vector sparsity and block sparsity. In terms of GPU acceleration, our implementation for balanced sparsity achieves around 1.4x-2.6x speedup compared to others.

\subsection{Discussions}

\paragraph{Visualization} We use the visualization method to understand why we can achieve a high accuracy close to random sparsity. 
Figure~\ref{fig:weights} shows a random-selected 64 $\times$ 64 block from the same position of 1500 $\times$ 1500 weight matrix in our LSTM experiment, under the sparsity ratio of 90\%.
The colored regions of the figure indicate non-zero parameters. 
Figure~\ref{fig:weights}c shows that, for block sparsity, the remaining blocks are randomly distributed, while intra-block, it is a dense weight matrix, suitable for parallel acceleration. 
After pruning, the weight map of \ourmethod{} is very similar to random sparsity. 
Thus, \ourmethod{} and random sparsity can maintain good accuracy. 
Besides, the visualization also indicates that \ourmethod{} is in a balanced weight distribution, compared with random sparsity, which provides a valuable feature for GPU inference acceleration.
In other words, each weight matrix row contains two blocks while each block contains three non-zero weights.

\begin{table}[t]
\centering
\resizebox{0.9\linewidth}{!}{
\begin{tabular}{c|c|ccc}
\hline
\multicolumn{2}{c|}{\multirow{2}{*}{Model}}                                                       & \multicolumn{3}{c}{Perplexity on Sparsity} \\ \cline{3-5} 
\multicolumn{2}{c|}{}                                                                             & 60\%         & 70\%         & 80\%         \\ \hline\hline
\multirow{3}{*}{\begin{tabular}[c]{@{}c@{}}Block\\ Sparsity\end{tabular}}    & block size: 4*4    & 80.6         & 83.2         & 88.1         \\
                                                                             & block size: 8*8    & 82.4         & 86.4         & 95.2         \\
                                                                             & block size: 16*16  & 83.7         & 88.3         & 99.5         \\ \hline\hline
\multirow{3}{*}{\begin{tabular}[c]{@{}c@{}}Balanced\\ Sparsity\end{tabular}} & balance range: 25  & 78.3         & 78.6         & 79.4         \\
                                                                             & balance range: 50  & 78.4         & 78.7         & 79.2         \\
                                                                             & balance range: 100 & 78.4         & 78.6         & 79.2         \\ \hline
\end{tabular}
}
\caption{Perplexity results on PTB dataset with different block size / balance range settings.}
\label{table:brange}
\end{table}

\paragraph{Sensitivity} We also study the sensitivity of our \ourmethod{} method by tuning the balance range. 
To show this more clearly, we take the block size of block sparsity as a comparison.
Table~\ref{table:brange} shows how the pruned model accuracy changes based on both different sparsity ratio and different balance ranges / block sizes. 
In this case, \ourmethod{} keeps the same model accuracy regardless of the change of balance range value. Even a very small balance range value (i.e. 25) cannot hurt the model accuracy.
On the contrary, for block sparsity, the light change of block size can lead to a significant perplexity increase.

\section{Conclusion} \label{conclusion}

In this work, we have proposed \ourmethod{}, a new fine-grained sparsity pattern to represent weight matrices in deep neural networks.
Experimental results on a set of neural networks show that \ourmethod{} achieves almost the same model accuracy as random sparsity with various sparsity ratios.  
Our measurements in widely-used deep learning workloads show that our efficient GPU implementation for \ourmethod{} can achieve significant speedup, up to 3.1x on GPU without accuracy loss. 
Our method shows not only the feasibility, but also the high potentials, for widely deployment of sparsity in neural network inference.

\section{Acknowledgements}
We would like to thank Dr. Ming Wu from Conflux and Dr. Yun Wang from Microsoft Research Asia for their valuable suggestions on improving this paper. We also thank the anonymous reviewers for their insightful feedbacks and comments. Shijie Cao was partly supported by National Nature Science Foundation of China (No.61772159).

\bibliography{refs.bib}
\bibliographystyle{aaai}
\end{document}